\documentclass[sigconf]{acmart}

\usepackage{graphicx} 
\usepackage{multirow}
\usepackage{multicol}
\usepackage{rotating}
\usepackage{amsmath}
\usepackage{amsthm}
\usepackage{amsfonts}
\usepackage{booktabs}
\usepackage{verbatim}
\usepackage{subfigure}
\usepackage{algorithm}  
\usepackage{algorithmic}
\usepackage{array}
\usepackage{bm}
\usepackage{url}
\usepackage{xspace}
\usepackage{enumitem}

\usepackage{makecell}

\newcommand{\cjy}[1]{{\color{black}#1}}
\newcommand{\Jeff}[1]{{\color{black}#1}}

\def\mW{{\bm{W}}}
\def\mQ{{\bm{Q}}}
\def\mK{{\bm{K}}}
\def\mV{{\bm{V}}}

\newcommand{\wq}{\mW_q}
\newcommand{\wk}{\mW_k}
\newcommand{\wv}{\mW_v}

\AtBeginDocument{%
  \providecommand\BibTeX{{%
    \normalfont B\kern-0.5em{\scshape i\kern-0.25em b}\kern-0.8em\TeX}}}

\setcopyright{acmcopyright}

\copyrightyear{2022} 
\acmYear{2022} 
\setcopyright{acmlicensed}\acmConference[IJCKG'22]{The 11th International Joint Conference on Knowledge Graphs}{October 27--29, 2022}{Virtual Event, China}
\begin{document}

\title{LaKo: Knowledge-driven Visual Question Answering via Late \\ Knowledge-to-Text Injection}

\author{Zhuo Chen}
\email{zhuo.chen@zju.edu.cn}
\affiliation{%
 \institution{College of Computer Science and Technology, 
  Zhejiang University}
\city{}
  \country{}
}

\author{Yufeng Huang}
\email{huangyufeng@zju.edu.cn}
\affiliation{%
  \institution{School of Software Technology, 
  Zhejiang University}
\city{}
  \country{}
}

\author{Jiaoyan Chen}
\email{jiaoyan.chen@cs.ox.ac.uk}
\affiliation{%
  \institution{ Department of Computer Science,
  University of Oxford}
\city{}
  \country{}
  }
  
\author{Yuxia Geng}
\email{gengyx@zju.edu.cn}
\affiliation{%
  \institution{College of Computer Science and Technology, 
  Zhejiang University}
\city{}
  \country{}
}

\author{Yin Fang}
\email{fangyin@zju.edu.cn}
\affiliation{%
 \institution{College of Computer Science and Technology, 
  Zhejiang University}
\city{}
  \country{}
}

\author{Jeff Z. Pan}
\email{j.z.pan@ed.ac.uk}
\affiliation{%
 \institution{School of Informatics, The University of Edinburgh}
\city{}
  \country{}
 }

 \author{Ningyu Zhang}
\email{zhangningyu@zju.edu.cn}
\affiliation{%
\institution{School of Software Technology, Zhejiang University}
\city{}
  \country{}
}

\author{Wen Zhang}
\authornote{Corresponding author}
\email{zhang.wen@zju.edu.cn}
\affiliation{%
\institution{School of Software Technology, Zhejiang University}
\city{}
  \country{}
}

%
%
%


\begin{abstract}
Visual question answering (VQA) often requires an understanding of visual concepts and language semantics, \cjy{which relies on} external knowledge. 
\cjy{Most existing methods exploit pre-trained language models or/and unstructured text, but the knowledge in these resources are often incomplete and noisy.
Some other methods prefer to use knowledge graphs (KGs) which often have intensive structured knowledge, but the research is still quite preliminary.}
In this paper,  we propose 
\textbf{LaKo}, a \cjy{knowledge-driven} VQA method via \textbf{La}te \textbf{K}nowledge-to-text Injecti\textbf{o}n. 
\cjy{To effectively incorporate an external KG, we transfer triples into textual format and propose a late  injection mechanism for knowledge fusion.}
Finally we address VQA as a text generation task with an effective encoder-decoder paradigm, which achieves state-of-the-art results on OKVQA datasets.
\end{abstract}

\graphicspath{ {figures/} }




\maketitle

\section{Introduction}
\Jeff{The task of} Visual Question Answering (VQA) \citep{antol2015vqa} is to answer natural language questions according to given images.
Recently, some VQA methods \citep{DBLP:journals/pami/WangWSDH18,DBLP:conf/cvpr/MarinoRFM19,DBLP:conf/aaai/ShahMYT19} are developed to utilize the external knowledge for open-world scene understanding (a.k.a. knowledge-based VQA). 
According to \Jeff{how to incorporate knowledge}, we divide the current works into two categories.

The first category is directly exploiting the knowledge in language model's parameters to answer questions \citep{DBLP:conf/emnlp/PetroniRRLBWM19,DBLP:conf/akbc/PetroniLPRWM020,DBLP:conf/emnlp/RobertsRS20,DBLP:conf/aaai/BianH0021}. 
Specifically, inspired by the knowledge-based language model in NLP field, some methods 
\citep{DBLP:journals/corr/abs-2101-06013,DBLP:conf/emnlp/GarderesZAL20,DBLP:conf/cvpr/MarinoCP0R21}  trying to \Jeff{inject} the common-sense or factual \cjy{knowledge}  \Jeff{as} part of the model's parameter \Jeff{during} model training. 
However,  the knowledge in language model sometimes is insufficient for VQA scenario, and they are likely to fail when referring to \textbf{new knowledge that is out of origin training corpus}. More importantly, these encoder-based finetuning frameworks with a MLP attached behind the last layer limits the utilization of knowledge within the model itself \citep{DBLP:conf/acl/GaoFC20}.

Works from the second category are based on the knowledge retrieval strategy.
We observe that those methods \citep{DBLP:conf/ijcai/ZhuYWS0W20,DBLP:journals/corr/abs-2109-04014,DBLP:conf/sigir/QuZ0CL21,DBLP:conf/sigir/JainKKJRC21} usually pass the vision-linguistic information through a search engine 
where the \textbf{network delay} might become a bottleneck. 
Others retrieve relevant corpus from encyclopedia articles, which leads to lots of \textbf{irrelevant information} and interferes with the model's judgment.

To address these challenges, in this paper, we propose \textbf{LaKo}, a knowledge driven VQA method via \textbf{La}te \textbf{K}nowledge-to-text Injecti\textbf{o}n, to effectively incorporate the external information, as shown in \cjy{Figure} \ref{fig:architecture}.
Specifically, we develop a \Jeff{retriever-reader VQA architecture, with} \cjy{a} 
knowledge retriever and a late injection mechanism between the knowledge and input corpus in reader.
First of all, 
we construct a common-sense knowledge graph (KG) for VQA, and 
the retriever queries this KG according to the vision-language input to recall the target triples. 
Specifically, the modality is unified  via the transformation of images to captions and triples to sentences with  Knowledge-to-Text strategy. Special prefixes are added to the front of them as knowledge guidance. 
Besides, we convert the VQA into a text generation task via an encoder-decoder paradigm, separate knowledge from the input corpus during encoding and integrate them at the stage of answer generation within decoder.
We observe that our method further boost the performance compared to traditional pipeline and obtain state-of-the-art results. 
To sum up, the main contributions are summarized below:
\begin{figure*}[htbp]
  \includegraphics[trim=0 0 50 0, width = 0.9\linewidth]{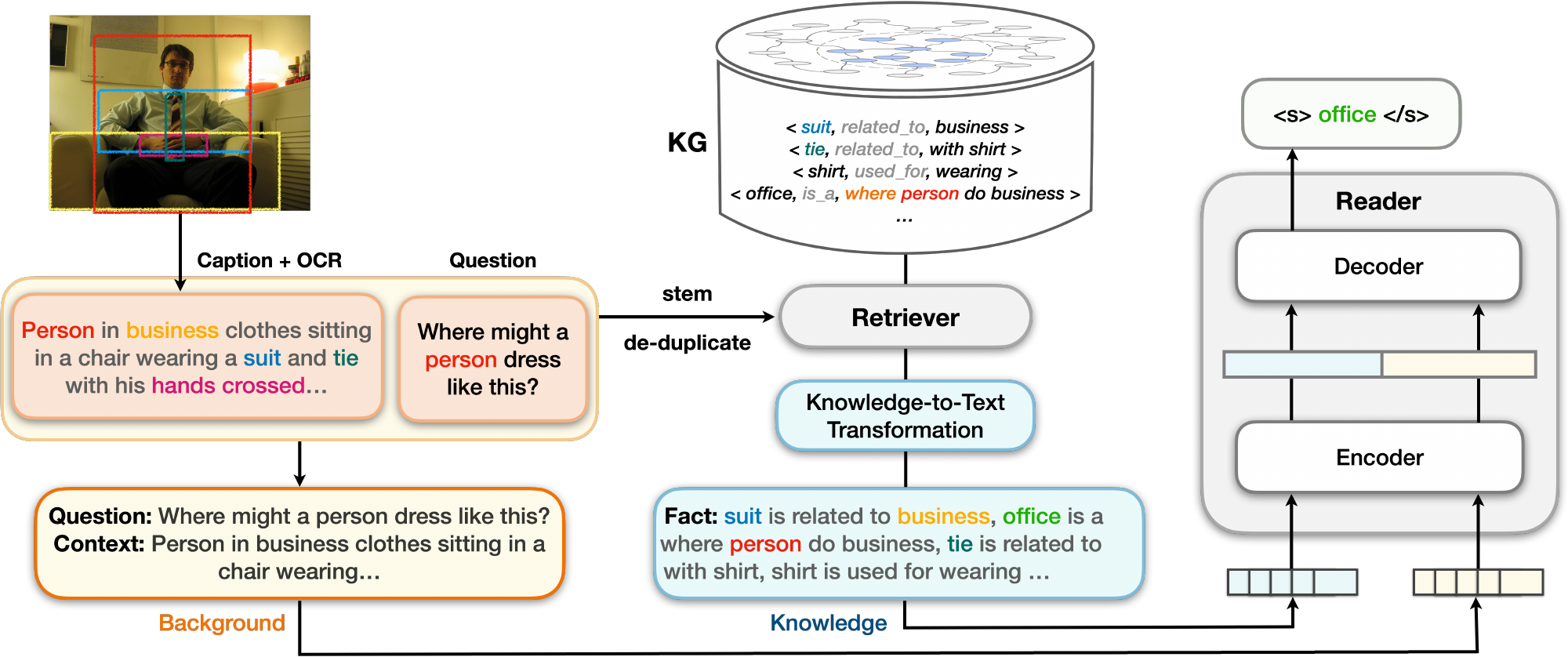}
    \caption{The overview model architecture of LaKo. Given a ($v,q$) pair, the generated background text and knowledge text retrieved by vision-language KG retriever are separately send to reader  for late knowledge injection. The predicted $ans$ tokens are decoded in turn.} 
  \label{fig:architecture}
  \vspace{-1mm}
\end{figure*}
\begin{itemize}
    \item We propose a new KG retrieval paradigm \Jeff{for} VQA together with a late knowledge injection strategy, which \Jeff{works without relying} 
    on annotated ground-truth knowledge.
    \item We improve and build a large-scale common-sense KG targeted at knowledge-based VQA, proving that high quality KG benefits VQA performance.
    \item Our method obtains state-of-the-art results on \cjy{the} OKVQA dataset, 
    \cjy{and verifies that using a high quality KG as the external knowledge is better than using unstructured text and pure language model parameters}. 
    Our code is available at \url{https://github.com/hackerchenzhuo/LaKo}.
\end{itemize}

\section{Related Work}
\subsection{Visual Question Answering (VQA)}
Since being proposed by \citep{antol2015vqa}, extensive VQA methods \citep{DBLP:conf/cvpr/00010BT0GZ18,DBLP:conf/nips/KimJZ18,DBLP:conf/cvpr/TeneyAHH18,DBLP:conf/semweb/0007CGPYC21} have emerged
to focus on applying multi-modal feature fusion between questions and images for answer decision. Recently, with the development of the pretraining technique, more  systems \citep{DBLP:conf/nips/LuBPL19,DBLP:conf/emnlp/TanB19,DBLP:conf/aaai/0010TYSTW021,DBLP:conf/eccv/Li0LZHZWH0WCG20,DBLP:conf/cvpr/ZhangLHY0WCG21,DBLP:conf/acl/LiGNXLL0020} begin to utilize multi-modal transformer architectures for the VQA task via vision-language pretraining (VLP) technique. 
\subsection{Knowledge-based VQA}
The knowledge-based VQA \citep{DBLP:journals/pami/WangWSDH18,DBLP:conf/cvpr/MarinoRFM19,DBLP:conf/aaai/ShahMYT19} requires the model to acquire factual or common-sense knowledge outside the image for question answering. According to the way of knowledge incorporation, we divide the current works into two categories.

\noindent\textbf{Exploit Knowledge in Language Model.}
Recent efforts in NLP fields \citep{DBLP:conf/emnlp/PetroniRRLBWM19,DBLP:conf/akbc/PetroniLPRWM020,DBLP:conf/emnlp/RobertsRS20,DBLP:conf/aaai/BianH0021,DBLP:conf/icml/WangYMLBLMZZY22} emphasize the relational world knowledge contained in huge pre-trained  language models (PLMs). 
Inspired by these perspectives, 
many works \citep{DBLP:journals/corr/abs-2109-08029,DBLP:journals/corr/abs-2106-13884,DBLP:journals/corr/abs-2109-05014} directly apply the PLMs into VQA tasks, where the PLM plays a role for question and image understanding (a.k.a. reader).
They hold the view that the knowledge within the PLMs is sufficient to support knowledge-based multi-modal tasks despite no additional knowledge input, which makes the VQA become a machine reading comprehension (MRC) problem. 
Specifically, Salaberria et al. \citep{DBLP:journals/corr/abs-2109-08029} convert an image into a caption\cjy{, and} then feed it together with question into the 
\cjy{BERT} \citep{DBLP:conf/naacl/DevlinCLT19} to predict the answer with an added classification head.
Nonetheless, the encoder-based finetuning framework with a MLP attached behind the {\tt [CLS]} position or pooling layer limits the utilization of the knowledge within the model itself \citep{DBLP:conf/acl/GaoFC20}.
Recent works \citep{DBLP:journals/corr/abs-2106-13884, DBLP:journals/corr/abs-2109-05014} endeavor to apply prompt for decoder-based model to solve the VQA problem under the few shot setting, but the max input length of model itself limits the knowledge utilization.

Some other works inject knowledge during the model training, letting the common-sense knowledge become part of the reader's parameters. For example, \citep{DBLP:journals/corr/abs-2101-06013} employ an auxiliary training objective that encourages the entity representation to align with the corresponding graph embedding in a KG.
ConceptBert \citep{DBLP:conf/emnlp/GarderesZAL20} introduces a multi-modal representation which learns a joint Concept-Vision-Language embedding. 
KRISP \citep{DBLP:conf/cvpr/MarinoCP0R21} exploits the implicit reasoning of transformer models, integrates symbolic representations from a knowledge graph (KG), and combines them together through a relational graph convolutional network (RGCN) \citep{DBLP:conf/esws/SchlichtkrullKB18}.

Although black-box models make knowledge an implicit representation, they are likely to fail when requiring implicit new knowledge that is out of the origin knowledge base (KB).
In this study, we not only make full use of the implicit knowledge in the PLMs, but also selectively carry out additional KG corpus to supplement those ancillary explicit knowledge which may be ambiguous within the PLMs. 
Besides, we decouple the knowledge from models, hoping that the reader could focus on understanding the input auxiliary corpus. This avoids the interference of model semantic understanding during the knowledge injection process, and keeps our model being sensitive to the new knowledge.


\noindent\textbf{Knowledge Retrieval Strategy.}
It is natural to think of adding a separate retrieval module (a.k.a. retriever) to recall the required explicit knowledge as external input of the downstream reader.
In order to take advantage of the information on the Internet, \citep{DBLP:conf/cvpr/MarinoRFM19,DBLP:journals/corr/abs-2109-04014,DBLP:conf/sigir/QuZ0CL21,DBLP:conf/sigir/JainKKJRC21} pass the vision-linguistic information through a search engine (e.g., Google) to retrieve relevant corpus (e.g., sentences from  \cjy{Wikipedia} articles or snippets in searching result) as weak positive knowledge samples, which are further passed to the reader module for knowledge incorporation.
Within the above methods, Luo et al. \citep{DBLP:journals/corr/abs-2109-04014} apply the previously retrieved snippets as a KB, and assign those snippets which contain the answer words as  weak-supervised signals for retriever training. 
Besides, Wu et al. \citep{DBLP:journals/corr/abs-2103-12248} leverage not only the structured knowledge, but also the image knowledge (from Google image search) to revise the answer.
However, the network delay might become a bottleneck for all these policies when take the search engine as the retriever, and the unstructured knowledge probably leads to the decrease of knowledge density.

Considering that highly dense knowledge is stored in structured KG triples,  
\citep{DBLP:journals/pami/WangWSDH18,DBLP:conf/nips/NarasimhanLS18,DBLP:conf/ijcai/ZhuYWS0W20} construct context-aware subgraph from a large scale KG (e.g., ConceptNet \cite{DBLP:conf/aaai/SpeerCH17}) 
based on entity name matching or embedding similarity.
But all of them preserve the original graph structure with a GNN-based model followed, which is deemed insufficient to exploit all useful evidence provided by external knowledge \citep{DBLP:conf/aaai/BianH0021}.
In this paper, we propose a vision-language KG retriever together with a Knowledge-to-Text transformation strategy. 
They unify the structured knowledge and visual data into a text modality to exploit the semantic understanding capability of PLMs, rather than relying on the message passing mechanism in GNNs. 
\footnote{KAT \citep{DBLP:conf/naacl/GuiWH0BG22}, K-LITE \citep{DBLP:journals/corr/abs-2204-09222}, TRiG \citep{DBLP:journals/corr/abs-2201-05299}, RA-VQA \citep{DBLP:journals/corr/abs-2210-03809}, REVIVE \citep{DBLP:journals/corr/abs-2206-01201}  and VLC-BERT \citep{DBLP:journals/corr/abs-2210-13626} are the works in the same period as ours so far, which all take advantage of the knowledge bases and the richness of PLMs.}

\section{Methodology}
 A VQA task is to provide an answer $ans$ given an image $v$ paired with a question $q$.
 Following \citep{antol2015vqa},  there are a list of (usually ten) acceptable ground truth answers ($GT_{ans}$) for each ($v$, $q$) pair.
% ...
\subsection{Vision-Language KG Retriever}
For the retrieval of factual triples from large scale KB, some works \citep{DBLP:conf/mm/Li0020,DBLP:conf/nips/NarasimhanLS18} utilize all detected objects in an image as the reference for knowledge retrieval. 
However, this brings in a lot of irrelevant noise and makes the model easy to lose focus, especially when the number of appeared objects is not small.
We observe that the image caption naturally contains a human-like attention mechanism on vision. 
So, instead of directly using vision modality data, our first step is to transform the visual content into the textual format.
Given a ($v$, $q$) pair, we convert the input image $v$ into corresponding image caption $C(v)$ with a pre-trained model at first. 
Besides, we further apply optical character recognition (OCR) technique \citep{DBLP:journals/corr/VeitMNMB16} for text extraction to improve the information integrity, and concatenate them with $C(v)$ to get an image representation $\tilde{v}$ with a text form:
$\tilde{v}=\mathcal{C}oncat[O(v);C(v)]$, where $O(.)$ denotes OCR output.
We note that  the pre-trained captioning model and the OCR model could be regarded as the modules for  caption feature extraction. Similar to ResNet \cite{DBLP:conf/cvpr/HeZRS16} for image feature or BERT \cite{DBLP:conf/naacl/DevlinCLT19} for textual feature, these models generate the simple descriptions toward the image  that are possibly related to ground truth background knowledge but not exactly equal to.

A KG may contain thousands of facts about a concept, but only several of them are relevant to the given ($v$, $q$).
Therefore, we reduce the scope of the KG through the establishment of a stem corpus in the VQA field, and make sure that all triples in the KG contain at least one stem within this corpus (see Section~\ref{sec:KGc} for details).
Then, we query the KG to get target triples based on a Knowledge-to-Text technique and a newly stem-based BM25 \citep{DBLP:journals/ftir/RobertsonZ09} algorithm.

\noindent\textbf{Knowledge-to-Text Transformation}. Knowledge facts are usually triples while the questions and answers are textual format. To realize the unification of three different modalities (i.e., vision, unstructured language, and structured knowledge) data, firstly, we translate the KG factual triples $t_{f}$ into  the sentence $s_{f}$. 
Specially, for those relations with high frequency, we follow the template-based method in \citep{DBLP:conf/aaai/BianH0021,DBLP:conf/eacl/HeinzerlingI21} with pre-defined cloze templates and conduct manual calibration.
For those long-tail or newly added relations, 
we apply BERT \citep{DBLP:conf/naacl/DevlinCLT19} tokenizer for coarse-grained word segmentation (e.g., ``locatedin'' is converted into ``located'' ``\#\#in'' ), and then generate cloze templates automatically after normalization.
Despite many relations are readable for the PLMs even without the above preprocess, other matching-based retrieval approaches like BM25 will benefit greatly from those templates.

\noindent\textbf{Stem-based BM25.}
The major discrepancy from the original BM25 is that our stem-based one defines the a word stem \footnote{ We get word stem via Porter Stemmer algorithm \url{https://tartarus.org/martin/PorterStemmer}} 
as the smallest semantic unit rather than an entire word. 
Our motivation is to maximize the knowledge from the limited VQA and KG corpus via stem merging.
Particularly, we remove these extra words with repeated stems in $\tilde{v}$ before concatenating it with stem in $q$ rather than de-duplicate on final $S_{query}$, since we want to maintain those important information emphasized by the $q$. 
Then we get a stem-based sequence $S_{query}$ with $s_1$,$s_2$,...,$s_t$ , and calculate the score for each factual triple sentence $s_{f}$ via: 
\begin{equation}
\operatorname{Score}(S_{query}, s_{f})=\sum_{i=1}^{t} w_{i} * R\left(s_{i}, s_{f}\right)\,,
\end{equation}
where $w_i$ represents the significance of $s_i$:
\begin{equation}
w_{i}=IDF\left(s_i\right)=\log \frac{N-n\left(s_i\right)+0.5}{n\left(s_i\right)+0.5}\,,
\end{equation}
where N denotes the total number of $s_{f}$ in KG and $n(s_i)$ denotes the number of $s_{f}$ containing the stem $s_i$. Hyperparameter $0.5$ is mainly for smoothing computation and $R(s_i,s_{f})$ measures the semantics correlation between $s_{i}$ and $s_{f}$ \citep{DBLP:journals/ftir/RobertsonZ09}.

Finally,  $s_{f}$ are retrieved according to their stem-based BM25 score. The Top-K $s_{f}$ are concatenated to get the $S_{fact}$  as external knowledge for each ($v$, $q$) pair, which contributes to the late knowledge injection within the reader.

\subsection{Late Knowledge Injection}
Recent VLP-based methods are substantially based on \cjy{an} encoder architecture  with a MLP attached \cjy{to} the {\tt [CLS]} position or the pooling layer, which limits the utilization of knowledge within the pre-trained model \citep{DBLP:conf/acl/GaoFC20}.
Inspired by previous works \citep{DBLP:conf/eacl/HeinzerlingI21,DBLP:conf/acl/CaoLHSYLXX20} \cjy{which explore the knowledge within PLMs}, we unify all data into textual to \cjy{fully} exploit the semantic understanding capability of the text-only PLM.
Specifically, we apply \cjy{the} encoder-decoder transformer architecture as the reader, following \citep{DBLP:conf/icml/ChoLTB21} to convert the knowledge-based VQA from a classification task into a text generation task.
Differently, considering that the entities in the KG do not exist in isolation and a closed loop is formed among triples, 
 we adapt the Fusion-in-Decoder (FiD) \citep{DBLP:conf/eacl/IzacardG21} into a new late injection paradigm to avoid the interference of vision-language information on knowledge self-integrate process.
 \begin{figure}[htbp]
 \vspace{-1mm}
  \includegraphics[trim=20 0 10 0, width = 0.65\linewidth]{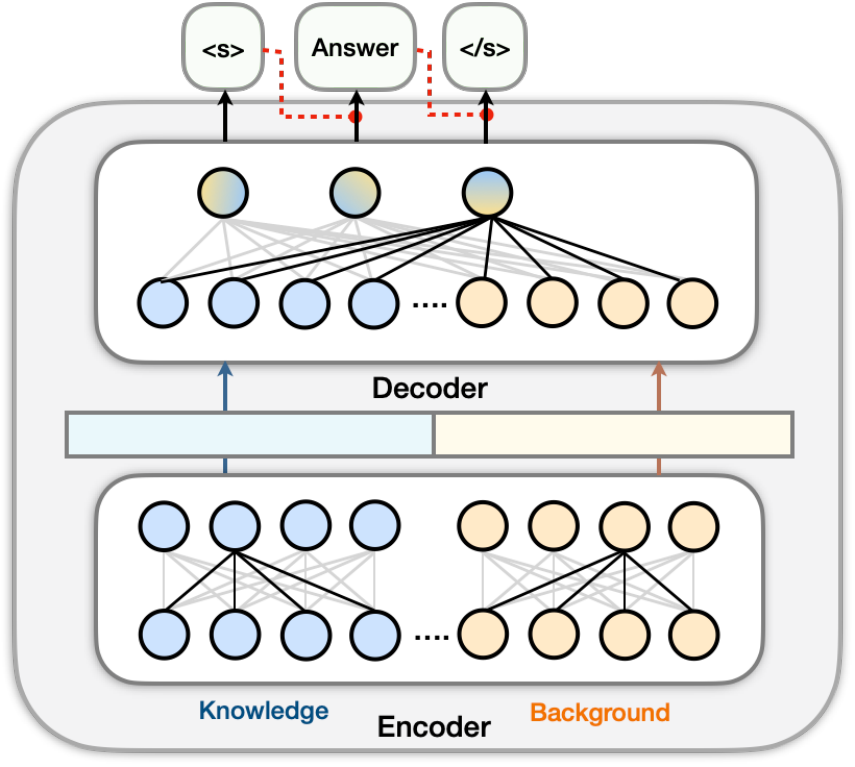}
  \vspace{-1mm}
    \caption{The self-attention architecture for late knowledge injection. The knowledge and background information are interact in encoder independently and fused in decoder to collectively predict the answer.
    } 
  \label{fig:late}
\end{figure}
As the self-attention architecture of Late Knowledge Injection shown in Figure \ref{fig:late}, this policy can achieve the independent encoding within  $S_{fact}$ for knowledge aggregation, and assist on late joint decoding among $(q,\tilde{v},S_{fact})$ for knowledge searching.
Besides, independent encoding also reduces the computation within the encoder during the self-attention process from $O((N+M)^2)$ to $O(N^2+M^2)$ where $N$ and $M$ denote the length of the knowledge and background input, respectively. 

In particular, we first add special prefixes  {\tt question:}, {\tt context:} and {\tt fact:} 
before the $q$, $\tilde{v}$ and $S_{fact}$ as the knowledge guidance, which make up two texts with independent semantics: 
\begin{itemize}
	\item Background context:  [{\tt question:} $q$, {\tt context:} $\tilde{v}$]
	\item Knowledge context: [{\tt fact:} $S_{fact}$]
\end{itemize}
Then the encoder independently processes the background and knowledge context through  $N$ layers transformer \citep{DBLP:conf/nips/VaswaniSPUJGKP17}. 
The output hidden state from each layer of the encoder form a \textbf{global representation} $X$ of dimension $(L_{b}+L_{k})*d$, where $L_{k}$ / $L_{b}$ denote the length of tokenized knowledge / background context, and $d$ is the  hidden state dimension of the model. 
As a regular autoregressive model, there are self-attention, cross-attention and feed-forward modules in each layer. For each head in one transformer layer of the decoder, the attention is defined as follow:
\begin{equation}
	\mathrm{Attn}(\mQ, \mK, \mV) =\operatorname{softmax}\left(\frac{\mQ\mK^{T}}{\sqrt{d}}\right) \mV\,,
\end{equation}
where $\mK$, $\mQ$, $\mV$ denotes matrices of key, query, value for input tokens \citep{DBLP:conf/nips/VaswaniSPUJGKP17}, respectively, and 
$\mQ=\wq H$, $\mK=\wk H$, $\mV=\wv H$ denotes the output of previous (self) attention layer ($H \in \mathbb{R}^{d}$).
Specially, the cross-attention process in each layer of the decoder is the only part for message exchange between encoder and decoder, where $\mK=\wk X$ and $\mV=\wv X$, which is the key of late knowledge injection.

Finally, the $ans$ tokens is decoded one by one with the begin of the start token (e.g., {\tt <s>}), and stop with the  end token (e.g., {\tt <$\backslash$s>}). Meanwhile, the whole encoder-decoder framework is optimized via minimizing the negative log-likelihood:
\begin{equation}
\mathcal{L}_{\theta}=-\sum_{j=1}^{|y|} \log P_{\theta}\left(y_{j} \mid y_{<j}, q, v, S_{fact}\right)\,,
\end{equation}
where $y$  are tokenized from $GT_{ans}$ for a given ($v$,$q$) pair.

\section{Experiments}
\subsection{Dataset}

\noindent\textbf{VQA2.0}
\citep{antol2015vqa} is a large standard VQA dataset containing about 1.1 million open-ended questions with 204,721 images. Each question is associated with 10 different answers obtained by \cjy{crowdsourcing}.

\noindent\textbf{OKVQA}
\cite{DBLP:conf/cvpr/MarinoRFM19} is a recent dataset where the visual content of \cjy{an} image is not sufficient to answer the question. There is not any exact ground truth common-sense fact triple for question support and all $ans$ are annotated by volunteers. In addition, all the images are from COCO 2014 validation set.

We note that other datasets like the FVQA \cite{DBLP:journals/pami/WangWSDH18} is smaller and easier than OKVQA since it targets at a reasoning over a given KB rather than visual reasoning with the open knowledge. Thus we mainly focus on OKVQA for model validation.

\subsection{Knowledge Graph Construction} \label{sec:KGc}
Following \citep{DBLP:conf/cvpr/MarinoCP0R21}, we  consider taking the common-sense knowledge (e.g., what are paper made of) and scientiﬁc knowledge (e.g., what genus are cats) to construct a new KG toward the knowledge-based VQA.
Differently, we exclude situational knowledge (e.g., where do cars tend to be located) outside since it may mislead the reader sometimes when the image scene is not typical (e.g., unseen scenarios).
Several knowledge sources are incorporated (we manually add ``$\_$'' on relations just for easy reading): 
\begin{itemize}
\item\textbf{ConceptNet} \citep{DBLP:conf/aaai/SpeerCH17} is a semantic network which contains human common-sense knowledge about the world;

\item\textbf{WebChild} \citep{DBLP:conf/acl/TandonMW17} contains triples which connect nouns with adjectives via more fine-grained relations (e.g., $``has\_shape"$, $``faster"$);

\item\textbf{DBpedia} \citep{DBLP:conf/semweb/AuerBKLCI07} \cjy{includes knowledge extracted from Wikipedia, which covers many fields and our daily life;}

\item\textbf{hasPart KB} \citep{DBLP:journals/corr/abs-2006-07510} collects $``has\_part"$ relationships between common objects such as <$dog$, $has\_part$, $whiskers$> or scientific ones like <$molecules$, $has\_part$, $atoms$>; 
\end{itemize}
Firstly, we collect triples from the above four knowledge sources to constitute the original KG (more than 900K triples). 
In particular, in  WebChild we filter the first 100K triples according to the normalized triple confidence score. The relations from Dbpedia are mainly $``category"$, and about 50k $``has\_part"$ triples are fetched form hasPart KB. Sevaral relations with a large number  but low potential contribution (e.g., $``Synonym"$, $``Antonym"$) are removed.

Next, we collect all of the symbolic entities from the dataset, including the words on questions, answers, generated image caption, and OCR recognition.
Based on the distribution of \textbf{word frequency} (from high to low), we remove common stop words and keep those that may have impact on knowledge representation (e.g. $``can"$). 
The \cjy{remaining words}
constitute the \textbf{VQA corpus} with their stem representation and we only retain those triples in KG whose heads and tails both contain stems in VQA corpus.

In addition, we define those relations \cjy{that occur more than 10,000 times in our KG (e.g. $``related\_to"$) as the frequent relations. }
\cjy{For the triples that have identical subjects and objects (e.g., <$person$, $related\_to$, $hand$> and <$person$, $has\_part$, $hand$>), 
we remove those triples associated with frequent relations}.
For example, according to our statistics, $13836$, $2584$, $2533$, $2391$ triples are deleted, which separately contain the relation of $``related\_to"$, $``used\_for"$, $``at\_location"$, $``is\_a"$.
Finally, we got \cjy{a KG with 300,559 triples, 96191 entities and 2198 relations.}
\subsection{Metrics} \label{sec:acc}
\noindent\textbf{$\mathbf{Acc}$.} For those classifier-based approaches, we apply the standard 
evaluation code\footnote{\url{https://github.com/GT-Vision-Lab/VQA}}, which 
\cjy{calculates the accuracy (Acc)} metric recommended in the VQA challenge \citep{antol2015vqa}:
\begin{equation}
	\operatorname{Acc}(ans)=\min (1, \frac{\#\{{ human~that~said~that~ans }\}}{3}). 
\end{equation}

\noindent\textbf{$\mathbf{EM}$.} For our text generation VQA framework, we use exact match (EM)  when calculate the Acc, where the generated $ans$ is compared to the ground truth answers ($GT_{ans}$) after normalization. 
Nevertheless, the unfixed answer length make\cjy{s} the size of answer space indefinite (i.e., even much bigger than \cjy{the PLM's vocabulary size}) for the autoregressive model. Likewise, the EM metric cannot make fair judgment sometimes which may result in some potential answers being left out (e.g., $GT_{ans}$ is ``in oven'' while $ans$ is ``oven'').
Hence, we introduce two \cjy{new metrics that are EM variants:}

\noindent\textbf{$\mathbf{Inc}$.} Inclusion-based Acc metric  regards the $ans$ as correct when it includes \cjy{one answer in $GT_{ans}$ or is included by one answer} in $GT_{ans}$ after normalization.

\noindent\textbf{$\mathbf{Stem}$.} Stem-based  Acc metric makes a small change toward $\mathbf{Inc}$: making judgment according to whether $ans$ and $GT_{ans}$ have an intersection on stem (e.g., the stem of $``happy"$ and $``happiness"$ are both$``happi"$).

It is noteworthy that all the $\mathbf{EM}$, $\mathbf{Inc}$ and $\mathbf{Stem}$ metrics will preferentially match the high score answer first. Meanwhile, the normalization procedure should strictly remove those stop words in both $ans$ and $GT_{ans}$ to avoid the disturbance.

\subsection{Training Details}
During knowledge retrieval, we utilize a SOTA caption model VinVL \citep{DBLP:conf/cvpr/ZhangLHY0WCG21} for \cjy{generating image descriptions, and we finetune its pre-trained checkpoint with COCO 2014 training set to prevent the leakage of $v$ in test data.
The top-10 generated factual sentence $s_f$ ($K$ = 10) are adopted to constitute the final $S_{fact}$.}
According to our statistics, the average word number in $S_{fact}$ is about $57$.

We respectively initialize the reader in LaKo-large/base with ofﬁcial pre-trained T5-large/base parameters.
We utilize AdamW optimizer with initialized learning rate 4e-5 (with warm-up ratio 6\%) and the model is trained for 20 epochs with max squence length 130, early stop patience 5 (i.e., the training algorithm waits 5 epochs before early stop if no progress on the validation set).
For the large version of LaKo, the mini-batch size is set as 8, which is 16 in the base version.

\subsection{Overall Results}
\begin{table*}[htbp]
    \centering
    \caption{Performance (\%) on the OKVQA test-split. 
    Since we only compared with those LM-based or VLP-based models, we do not highlight LM on ``Knowledge Src''.
    The full names of these abbreviations are as follows: GS (Google Search), W (Wikipedia), KG (Knowledge Graph), GI (Google Image), Enc.Dec.(Encoder-Decoder), TG (Text Generation), IE (Information Extraction), CLS (Classification), Src (source), Acc (Accuracy), C (Caption), T (Tag).
    }
	\vspace{-1mm}
    \resizebox{0.68\textwidth}{!}{
    \begin{tabular}{lcccc}
    \toprule
    {\bf Method}  & \makecell[c]{\bf Backbone} & \makecell[c]{\bf Architecture} & \makecell[c]{\bf Knowledge Src} & {\bf Acc} \\
    \toprule
    ConceptBert~\citep{DBLP:conf/emnlp/GarderesZAL20}  & BERT & Encoder (CLS) & KG & 33.66 \\
    KRISP~\citep{DBLP:conf/cvpr/MarinoCP0R21}  & VisualBERT & Encoder (CLS) & W \& KG  & 38.35 \\
    MAVEx~\citep{DBLP:journals/corr/abs-2103-12248}  & ViLBERT & Encoder (CLS) & W \& KG \& GI & 38.70 \\ 
    Caption-DPR + CReader~\citep{DBLP:journals/corr/abs-2109-04014} & LXMERT & Encoder (CLS) & GS & 36.78 \\ 
    Caption-DPR + EReader~\citep{DBLP:journals/corr/abs-2109-04014} & RoBERTa & Encoder (IE) & GS & 39.20 \\ 
    KGE Aligning~\citep{DBLP:journals/corr/abs-2101-06013} & LXMERT & Encoder (CLS) & KG & 39.04 \\ 
    CBM + MMBERT~\citep{DBLP:journals/corr/abs-2109-08029}  & BERT + MMF & Encoder (CLS) & - & 39.20 \\
    PICa-Base (C+T)~\citep{DBLP:journals/corr/abs-2109-05014} & GPT-3 & Decoder (TG) & GS & 43.30 \\ 
    \hline
    LXMERT \citep{DBLP:conf/emnlp/TanB19} & - & Encoder (CLS) & - & 36.91\\
    LXMERT \citep{DBLP:conf/emnlp/TanB19} + Knowledge & - & Encoder (CLS) & KG & 37.92\\
    T5\citep{DBLP:conf/emnlp/TanB19} + Prefixes & - & Enc.Dec. (TG) & - & 42.03\\
    \hline
    \textbf{LaKo} & T5 & Enc.Dec. (TG) & KG & \textbf{47.01}\\
    \bottomrule
    \end{tabular}
    }
        \label{tab:main-result}
    \vspace{-1mm}
\end{table*}
Table \ref{tab:main-result} summarizes the main result on the OKVQA dataset. 
Our LaKo follows the large version with late knowledge injection strategy and is re-pretrained using VQA2.0 training data unless otherwise specified.
Specifically, we compared \cjy{LaKo} with recent LM-based (BERT \cite{DBLP:conf/naacl/DevlinCLT19}, RoBERTa \cite{DBLP:journals/corr/abs-1907-11692}, GPT-3 \cite{DBLP:conf/nips/BrownMRSKDNSSAA20}) or VLP-based ( VisualBERT~\cite{DBLP:journals/corr/abs-1908-03557},  ViLBERT~\cite{DBLP:conf/nips/LuBPL19}, LXMERT~\cite{DBLP:conf/emnlp/TanB19}, MMF \cite{singh2020mmf}) protocols for fair\cjy{ness}.
In addition to these methods, we also \cjy{leverage some} other strong baselines: 
\begin{itemize}
	\item \textbf{LXMERT} \citep{DBLP:conf/emnlp/TanB19}. A SOTA two stream VLP model, where we extract image region features from \cjy{a} pre-trained Faster R-CNN \citep{DBLP:journals/pami/RenHG017} as visual input.
	\item \textbf{T5} \citep{DBLP:conf/emnlp/TanB19}. A recent seq-to-seq LM with encoder-decoder paradigm. We simpl\cjy{y} take the question and caption as the input with special prefixes put ahead\footnote{Other encoder-decoder models such as BART \cite{DBLP:conf/acl/LewisLGGMLSZ20} performs not as good as T5 as we have tested in VQA field. Thus we take T5 as the backbone of LaKo.}.
\end{itemize}

We find that many previous works involve explicit knowledge retrieval from search engine\cjy{s} (e.g., ``Google Search'' and ``Google Image''), or store \cjy{a} huge number of unstructured encyclopedia texts as the background KB in advance (e.g.,``Wikipedia'').
However, in realistic application, the network bandwidth may become the bottleneck, while noise contained in unstructured text may limit the knowledge scale due to the limitation of max sequence length.
Nevertheless, we compare 
LaKo with these approaches, observing that the performance of our model still exceeds the SOTA method (PICa-Base (C+T), with 175B parameters GPT-3 without multi-query ensemble) and further improves by 3.71\%. 
We also concatenate the $S_{fact}$ with $q$ as input text to LXMERT, which achieves an improvement of 1.01\% on Acc.
Most importantly, since our results are based on EM, the actual acceptable answer accuracy may be higher than our statements. 
We believe that T5's generalization on VQA tasks mainly comes from its large pre-trained  Colossal Clean Crawled Corpus (C4) \citep{DBLP:conf/emnlp/TanB19} dataset , and the basic encoder-decoder architecture makes it flexible when migrated to other tasks. 
Moreover, we get rid of the dependence on search engines and annotated ground-truth knowledge, which is easy for other researchers to follow.

We guess that two points are the key issues to the low performance of those traditional classifier-based approaches:

\emph{(i)} In order to mitigate the long-tail problem in answers and the impact of disjoint answers between training $\&$ testing set (e.g., the total number of answers in VQA2.0 is 29,140 in our statistics, with just 15,259 intersections),
they have to artificially prescribe the answer candidate set based on occurrence (Occ.) frequency and therefore determine the output dimension of the last MLP layer in classifier. It is a \textbf{trade-off between answer coverage and error rate}. 
This also inevitably affect the upper bound of model performance: the oracle Acc of VQA2.0 drops to 92.86\% with Occ. 9, and OKVQA respectively drops to 72.08\% (Occ. 10), 85.38\% (Occ. 5), 91.44\% (Occ. 3).
\emph{(ii)} The pluggable MLP layer attached behind the PLM limits the direct utilization of the knowledge within the model, which is proved by many previous works like \citep{DBLP:conf/acl/GaoFC20}. 

Besides, the phenomenon that recent SOTA works are rarely based on GNN supports the view that those GNN-based models may not fully exploit all evidence provided by external knowledge \citep{DBLP:conf/aaai/BianH0021} and the knowledge within the PLM is essential. Since the single-modal data is much richer compared to limited image-text paired data, it makes the text-only PLM even stronger than VLP model on knowledge-based VQA.
Furthermore, we hold the review that our model is not trained on natural \textbf{long language sequences} in the VQA field since the answer's average length is 1.3/1.2 words in OKVQA/VQA2.0. According to our statistics, the length of the answers generated by LaKo is 1.23, which supports the idea that making VQA a simpler generative task is beneficial. 


\subsection{Ablation Study} \label{sec:ablation}

\noindent\textbf{Effect of Knowledge and Finetuning.} In Table \ref{tab:Ablation of KF} we discuss the \cjy{impact} of factual knowledge and the pretraining on LaKo. 
``Finetune'' refers to re-pretraining LaKo on VQA2.0 before finetuning on OKVQA, which is necessary since the VQA is a relatively unfamiliar logic for text-only PLMs. 
We observe that the knowledge retrieval and the re-pretraining in LaKo-large lead to 2.71\% and 2.03\% improvement, respectively, which is constant in LaKo-base. We own this to the fact that massive implicit knowledge is contained in PLMs' parameter after the pretraining process on Internet corpus, and LaKo could further exploit those explicit external knowledge to support more precise answer prediction.
Besides, the result on Inc-based and Stem-based Acc usually can be 5\%-7\% higher than the original EM-based one, which shows the potential of our architecture toward multi-modal knowledge tasks.
\begin{table}[htbp]
\vspace{-1mm}
    \centering
     \caption{Ablation study (\%) for the effect of knowledge and finetuning. The full names of these abbreviations: w/o (without), KGR (Knowledge Graph Retrieval). ``Finetune'' refers to re-pretraining LaKo on VQA2.0 dataset first. 
    }
    \addtolength\tabcolsep{1pt}
    \resizebox{0.85\linewidth}{!}{
    \begin{tabular}{lccc}
    \toprule
    {\bf Method}  & \makecell[c]{\bf EM}  & \makecell[c]{\bf Inc} & {\bf Stem} \\
    \toprule
	LaKo-large & \textbf{47.01} & \textbf{53.09}  & \textbf{53.97}\\
	- w/o \{KGR\} & 44.74 & 50.90  & 51.70\\
	- w/o \{Finetune\} & 44.06 & 49.48  & 50.25\\
	- w/o \{Finetune \& KGR\} & 42.03 & 47.34  & 48.11\\
    \hline
	LaKo-base & 42.21 & 48.17  & 49.06\\
	- w/o \{KGR\} & 40.27 & 45.91  & 46.85\\
	- w/o \{Finetune\} & 39.89 & 45.05  & 45.93\\
	- w/o \{Finetune \& KGR\} & 38.71 & 43.80  & 44.32\\
    \bottomrule
    
    \end{tabular}
    }

    \label{tab:Ablation of KF}
    \vspace{-2mm}
\end{table}
\begin{table}[htbp]
\vspace{-2mm}
    \centering
    \caption{Result (\%) on \cjy{VQA}2.0 test-dev. 
    }
    \vspace{-1mm}
    \addtolength\tabcolsep{2pt}
    \resizebox{0.55\linewidth}{!}{
    \begin{tabular}{lcc}
    \toprule
    {\bf Method}  & \makecell[c]{\bf Acc}  & \makecell[c]{\bf Inc.} \\
    \toprule
	T5 & 66.38 & 68.85  \\
	ViLBERT \citep{DBLP:conf/nips/LuBPL19} & 67.90 & - \\
	LXMERT & 69.15 & -  \\
	LaKo & 68.07 & 70.47 \\
	VinVL \citep{DBLP:conf/cvpr/ZhangLHY0WCG21} & \textbf{75.95} & - \\
    \bottomrule
    \end{tabular}
    }

    \label{tab: Ablation of vqa}
    \vspace{-1mm}
\end{table}
\begin{table*}[htbp]
    \centering
    \caption{The performance (\%) discrepancy between late and early injection. }
    \vspace{-1mm}
    \addtolength\tabcolsep{2pt}
    \resizebox{0.85\textwidth}{!}{
    \begin{tabular}{lcccccc}
    \toprule
    \multirow{2}{*}{\bf Method}   
    ~& \multicolumn{3}{c}{LaKo-large} & \multicolumn{3}{c}{LaKo-base} \\
    \cmidrule(lr){2-4}\cmidrule(lr){5-7}
    ~ &  EM & Inc. &  Stem & EM & Inc. &  Stem \\
    \toprule
     Early Injection  & 46.37 &  52.26 & 53.29 & 41.41 & 47.54 &  48.22 \\
Late Injection & 47.01 ({\color{red}{$\uparrow$ 0.64}}) & 53.09 ({\color{red}{$\uparrow$ 0.83}}) & 53.97 ({\color{red}{$\uparrow$ 0.68}})& 42.21 ({\color{red}{$\uparrow$ 0.80}}) & 48.17 ({\color{red}{$\uparrow$ 0.63}})& 49.06 ({\color{red}{$\uparrow$ 0.84}}) \\
    \toprule
    \end{tabular}
    }
    
    \label{tab:one-or-two-stream}
\end{table*}

\begin{figure*}[htbp]
 \vspace{-1mm}
  \includegraphics[trim=0 0 20 0, width = 0.99\linewidth]{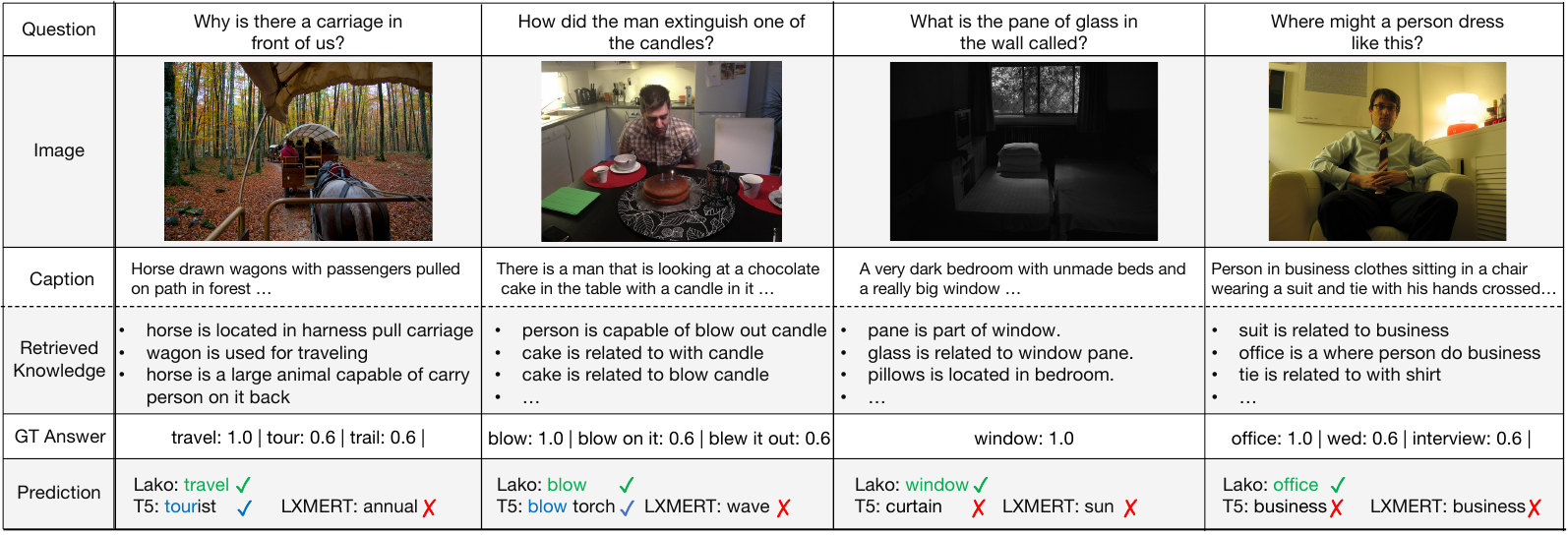}
    \caption{We visualize some predictions and their corresponding retrieved facts $s_f$. Special, each question is associated  with  10  different  answers, and the score of $GT_{ans}$ is calculated via $\min (1, \frac{\#\{{ human~that~said~that~ans }\}}{3})$. } 
  \label{fig:case}
  \vspace{-1mm}
\end{figure*}

Furthermore, we also make experiments on the standard \cjy{VQA}2.0 dataset as shown in Table \ref{tab: Ablation of vqa}, where the results reflect that the knowledge is also effective for generic VQA (\cjy{Acc is improved} from 66.38\% to 68.07\%), and LaKo is comparable to part of VLP models like ViLBERT \citep{DBLP:conf/nips/LuBPL19} and LXMERT \citep{DBLP:conf/emnlp/TanB19}. 
However, it is not easy for us to surpass those SOTA VLP model without large-scale VLP model pretraining and fine-grained image feature providing. 

\noindent\textbf{Beneﬁts of Late Injection.} 
We studied the impact of late knowledge injection, 
which separate the background text and knowledge text on input and focuses on the knowledge aware answer generation within decoding stage via  computing cross-attention between  current generative tokens and these input corpora. 
\cjy{According to} the results shown in Table \ref{tab:one-or-two-stream}, we can see that the late injection process has a stable improvement ($0.6$\%$\sim0.9$\%) on answer prediction, which suggests that LaKo 
\cjy{benefits} from late knowledge injection rather than simply put\cjy{s} all these texts together as the input. 

\noindent\textbf{Influence of KG quality.}
We discuss the influence of KG quality on final performance of LaKo.
As the result shown in Table \ref{lab:KG_quality}, since the stem limitation of VQA corpus offers effective constraint for the knowledge range of the model, it mitigates the distraction from irrelevant information.
Meanwhile, we observe that removing part of triples with redundant frequent relations also has a positive impact on the model.
\begin{table}[htbp]
    \centering
    \caption{``Stem Filter'': filtering out triples whose heads or tails do not contain stems in VQA corpus.
    ``Freq. Rm'': selectively removing those triples which contain frequent relations (see Sec. \ref{sec:KGc} for details).}
    \vspace{-1mm}
    \resizebox{0.92\linewidth}{!}{
    \begin{tabular}{lccc}
    \toprule
    {\bf KG}  & \makecell[c]{\bf EM}  & \makecell[c]{\bf Inc.} & {\bf Stem} \\
    \toprule
	Final version & \textbf{47.01} & \textbf{53.09}  & \textbf{53.97}\\
	- w/o \{Freq. Rm\} & 46.79 & 52.66  & 53.52\\
	- w/o \{Stem Filter \& Freq. Rm\} & 46.26 & 52.17  & 52.91\\
    \bottomrule
    \end{tabular}
    }

    \label{lab:KG_quality}
    \vspace{-1mm}
\end{table}

 \vspace{-1mm}
\subsection{Interpretability}\label{fig:interpretability}
\cjy{To demonstrate the effectiveness of LaKo and that the knowledge injection is effective,
we visualize some predictions and their corresponding retrieved facts $s_f$.
As illustrated in Figure \ref{fig:case}, LXMERT  outputs wrong predictions in all these four cases, and sometimes outputs totally unrelated answers (e.g., ``annual'' in the first case).
T5 outputs reasonable answers in the first and second cases, but makes mistakes in the third and fourth cases where there are distracting information within the caption or the query is ambiguous.
In contrast, some retrieved $s_f$ in LaKo are relevant to the ground truth answers (e.g, ``office is where person do business'' in the fourth case), which helps LaKo make correct prediction in those cases.
}
\begin{figure}[htbp]
  \includegraphics[trim=20 0 10 0, width = 0.9\linewidth]{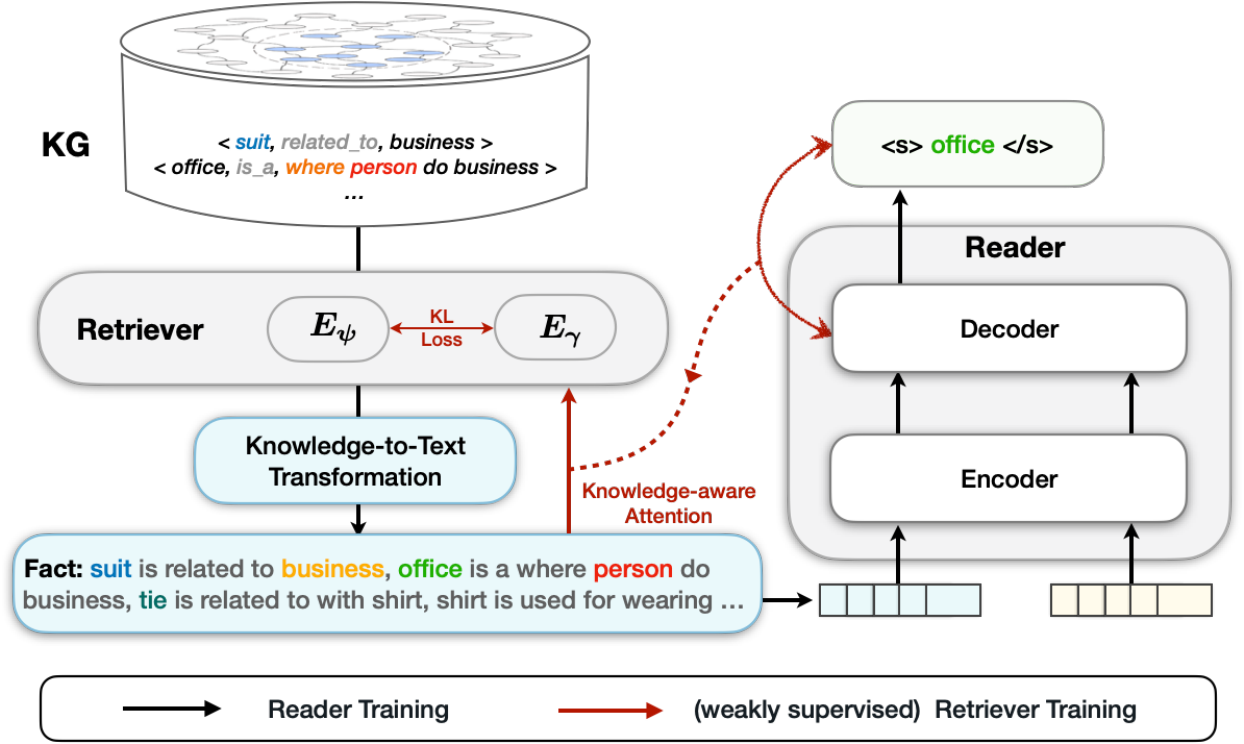}
    \caption{The training process of weakly supervised differentiable retriever.
    } 
  \label{fig:retriever}
  \vspace{-2mm}
\end{figure}

\subsection{Feasibility of Differentiable KG Retriever}\label{sec:differentiable_retriever}
\cjy{The KG retriever} plays an important role in our framework. So, \cjy{is it feasible to train the retriever for better performance?}
Following \citep{DBLP:conf/iclr/IzacardG21}, we leverage the cross attention between the token of the prediction output and input $S_{fact}$ for retriever training, as shown in Figure \ref{fig:retriever}. 
In particular, the minimum unit of attention score here is $s_f$. Therefore, no more than $K$ (i.e., 10 in our work) scores are generated for each ($v,q$) pair, which are then utilized to training the retriever as the weak-supervised signals of corresponding $s_f$. 
We claim the retriever as a pseudo-siamese network \cite{DBLP:conf/cvpr/ChenH21} with $E_{\gamma}$ for encoding $s_f$ and $E_{\psi}$  for encoding $S_{query}$, which only share the model architecture (BERT-base) rather than sharing parameters.
It is optimized through minimizing the $KL$-divergence:
\begin{equation}
\mathcal{L}_{\mathrm{KL}}=
\sum_{f \in \mathcal{K}_{kg}} {A}_{q, f}\left(\log {A}_{q, f}-\log \mathcal{O}(S_{query}, f)\right)\,,
\end{equation}
where $\mathcal{K}_{kg}$ denotes the collection of those (top-k) retrieved textual triples  via Knowledge-to-text transformation and 

\begin{equation}
    {A}_{q, f}=\frac{\exp \left(Atten_{q, f}\right)}{\sum_{f^{\prime} \in \mathcal{K}_{kg}} \exp \left(Atten_{q, f^{\prime}}\right)}\,,
\end{equation}
\begin{equation}
\mathcal{O}(S_{query}, f)=\frac{\exp \left(E_{\psi}(S_{query})^{T} E_{\gamma}(f)\right)}{\sum_{f^{\prime} \in \mathcal{K}_{kg}} \exp \left(E_{\psi}(S_{query})^{T} E_{\gamma}(f^{\prime})\right)}\,.
\end{equation}

For the aggregation of $Atten_{q, f}$, we apply several strategies: 

\begin{enumerate}

\item [1)] taking the max or average or the top $1/2$ attention scores over the input tokens corresponding to a $s_f$; 
\item [2)] taking the scores from last half layers or all the layers;
\item [3)] \cjy{adding} additional score bias (e.g., 1) to the $s_f$ which contained answer stems.

\end{enumerate}

\begin{figure}[htbp]
  \includegraphics[trim=20 0 10 0, width = 0.99\linewidth]{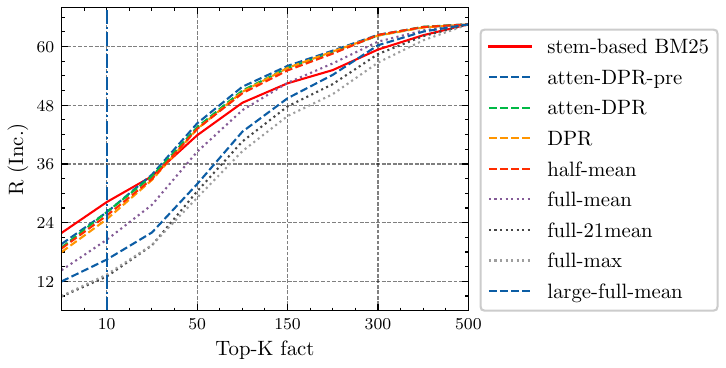}
  \vspace{-1mm}
    \caption{The results of recall \cjy{(R) by different retrievers.}
    Full names of these abbreviations: 
    full/half (taking the scores from all/last half of the layers), max/mean/21mean (taking the max/average/the top $1/2$ scores over the input tokens corresponding to a $s_f$), pre ( use the model finetuned at VQA2.0). Besides, DPR \citep{DBLP:conf/emnlp/KarpukhinOMLWEC20} refers to adding 1 additional score bias to the $s_f$ which contains the answer stems.
    } 
  \vspace{-2mm}
  \label{fig:inc}
\end{figure}

The motivation for choosing the top $1/2$ attention scores is to extract the attention signals from the most valuable part of each $s_f$ rather than a single token or simple averaging all the tokens;
Specifically, all the attention scores are computed toward \cjy{the} first output token of the decoder, and the scope of retrieval is within the TOP-500 fact\cjy{s} retrieved via stem-based BM25 (we also strive to retrieve from the whole KG with 300K triples via faiss engine \cite{johnson2019billion}, only to get a poor result with Recall 23.37\% on Top-100 facts). 
In order to measure the performance for retrieval, we introduce Inc-based Recall and define it \textbf{as a success recall when the retrieved knowledge sentence include the answer stem}. 
We train our retriever with different attention scores from both LaKo-large/base.

The Inc-based Recall rate for Top-K fact\cjy{s} are shown in Figure \ref{fig:inc}.
As we can see, the Recall rate
\cjy{by the stem-based BM25 is higher than the trained retriever\cjy{s} when $K$ is smaller than $20$, but becomes lower than some trained retrievers as $K$ increases.}
Since we only utilize the Top-10 $s_f$ for vision-language reading,  this result is not optimistic enough for iterative training on retriever, and we observe that the VQA Acc simultaneously drops with lower fact Recall rate.
We guess that multiple different entities within the caption/question and the sparse embedding space impact the  final similarity-based embedding recall, so we simply select the stem-based retrieval strategy in our primary experiment.

\section{Discussion and future work}
\begin{itemize}
	\item Given a ($v$, $q$) pair, sometimes there are several correct answers in real world, but the  generation-based approach only gets one answer at a time instead of sorting those potential ones. It is  an interesting direction to consider generating multiple answers with particular policy, where the trie-based search \cite{cormen2022introduction} strategy could be considered for generative answers' domain constrain over the candidates.
	\item KG driven zero-shot problem \cite{DBLP:conf/ijcai/ChenG0HPC21,DBLP:journals/corr/abs-2112-10006,DBLP:journals/corr/abs-2106-15047,DBLP:conf/www/GengC0PYYJC21} on VQA also deserves a deeper research to further discuss the trade-off between PLMs and KGs, which requires the model to have better generalization ability toward the real world scenarios.
	\item we believe that the differentiable KG retriever in VQA field would be practicable in the future with  better knowledge representation learning methods and high-quality knowledge annotations published. 
\end{itemize}
 
\section{Conclusion}
In this paper, we propose LaKo,
a knowledge-driven VQA method via late knowledge-to-text injection, to effectively incorporate both the knowledge from the KG and the PLM itself.
Specifically, we address the VQA as a text generation task with an effective encoder-decoder paradigm under vision-language retriever-reader architecture, which achieves state-of-the-art result on standard knowledge-based VQA dataset OKVQA.
Besides, we also pay attention to the KG construction, observing that KG with higher quality contributes to better performance of LaKo. This could be an exploration direction for future  works.
More importantly, we get rid of the dependence on annotated ground-truth knowledge and search engines, which is easy for other researchers to follow.

\section{Acknowledgements}
This work is funded by NSFCU19B2027/91846204 and the EPSRC project ConCur (EP/V050869/1).

\bibliographystyle{ACM-Reference-Format}
\bibliography{sample-base}

\end{document}